\pdfoutput=1

\documentclass[11pt]{article}

\usepackage[preprint]{acl}

\usepackage{times}
\usepackage{latexsym}
\usepackage{amsmath}
\usepackage{float}
\usepackage{multirow}
\usepackage{bbm}
\usepackage{booktabs}
\usepackage[T1]{fontenc}

\usepackage[utf8]{inputenc}

\usepackage{microtype}

\usepackage{inconsolata}

\usepackage{graphicx}
\usepackage{mdframed}       
\title{I Need Help! Evaluating LLM's Ability to Ask for Users' Support:\\A Case Study on Text-to-SQL Generation}

\author{%
    Cheng-Kuang Wu$^{1,2}$\thanks{Equal contribution},
    Zhi Rui Tam$^1$\footnotemark[1],
    Chao-Chung Wu$^1$,
    Chieh-Yen Lin$^1$,\\
    \textbf{Hung-yi Lee}$^2$\thanks{Equal advisorship},
    \textbf{Yun-Nung Chen}$^{2}$\footnotemark[2]\\
    $^1$Appier AI Research\\
    $^2$National Taiwan University \\
}

\begin{document}
\maketitle
\begin{abstract}
This study explores the proactive ability of LLMs to seek user support.
We propose metrics to evaluate the trade-off between performance improvements and user burden, and investigate whether LLMs can determine when to request help under varying information availability.
Our experiments show that without external feedback, many LLMs struggle to recognize their need for user support.
The findings highlight the importance of external signals and provide insights for future research on improving support-seeking strategies.
Source code: \url{https://github.com/appier-research/i-need-help}.
\end{abstract}

\section{Introduction}

The impressive instruction-following~\cite{wei2021finetuned} abilities of large language models (LLMs) have enabled their out-of-the-box usage to solve problems.
However, these models generate hallucinated content~\cite{rawte2023survey} or incorrect predictions in their efforts to fulfill user instructions, which undermines their reliability.

When LLMs generate incorrect outputs for a given instruction, the issue can be examined from multiple perspectives.
One is that the model simply lacks the \textit{competence} to satisfy the instruction, suggesting a straightforward solution: enhancing the model's capabilities, which is the focus of most previous research.
Another is that the model could actually solve the task with additional \textit{support}.
For instance, \citet{pourreza2023evaluating} found that models often fail due to underspecified natural language queries.
Similarly, \citet{li2024can} showed that while GPT-4 struggles initially, its performance can improve by up to 20.01\% with human-annotated external knowledge.
In such cases, models should proactively seek help rather than attempting to satisfy instructions with insufficient information.

Motivated by these considerations, we aim to investigate whether LLMs can identify when to ask for user support. Since providing such support requires additional effort from users, there is an inherent trade-off between ``LLM performance improvement from user support'' and ``user burden''. Therefore, we seek to answer the following research questions:
\textit{\textbf{RQ1:} How can we design evaluation metrics to quantify this trade-off?}
\textit{\textbf{RQ2:} How effectively do LLMs manage this trade-off, and what strategies are effective in improving it?}

\begin{figure}[t!]
  \includegraphics[width=\columnwidth]{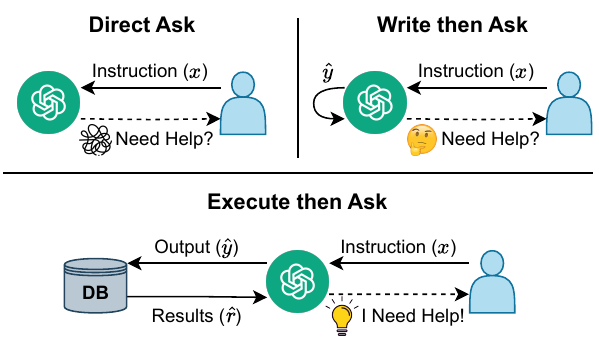}
  \caption{Overview of our experiments on text-to-SQL. LLMs struggle to determine when they need help based solely on the instruction ($x$) or their output ($\hat{y}$). They require external feedback, such as the execution results ($\hat{r}$) from the database, to outperform random baselines.}
  \label{fig:showcase}
\end{figure}

In this work, we focus on the text-to-SQL task as a case study to empirically investigate the aforementioned research questions.
We chose the text-to-SQL task for several reasons:
(1) Its promising applicability, empowering lay users to retrieve data with natural language queries.
(2) The inherent ambiguity in some natural language queries, leading to uncertainty in the generation of SQL code~\cite{pourreza2023evaluating}, making it suitable for scenarios where additional user support is beneficial.
(3) There exists a large-scale BIRD dataset~\cite{li2024can} with human-annotated external knowledge, providing a valuable source of user support for our empirical investigation.

\noindent Our contributions can be summarized as follows:

\begin{enumerate}

\item We propose metrics for evaluating the trade-off between performance improvement from user support and the associated user burden.

\item We conduct experiments using various methods to balance this trade-off, providing insights into LLMs' capabilities in seeking user support and identifying effective strategies for enhancing their performance.

\end{enumerate}


\section{Formulation for Seeking Support}
\subsection{General Setup}

Consider an LLM $f$ parameterized by $\theta$, along with a prompt template $p(\cdot)$.
Given a natural language instruction $x$, we use $z$ to represent \textit{support}, which should enhance the LLM's ability to fulfill $x$. Formally, $\hat{y}_z = f(p(x, z) \mid \theta)$ is more likely to satisfy $x$ compared to $\hat{y} = f(p(x) \mid \theta)$.
We denote the "ask for support" signal emitted by the LLM as $\hat{a}$, defined as a confidence score in the range $[0, 1]$, where $1$ indicates an absolute need for support.
A threshold $\tau$ is then used to determine whether to request $z$.
In practice, $\hat{a}$ could also be a natural language request specifying the \textit{type of support} needed by the LLM, which we leave for future work.

\subsection{Evaluation}
\label{sec:eval}
To measure the trade-off between \textit{performance improvement from user support} and \textit{user burden}, we need 2-dimensional evaluation.
One dimension is the user burden ($B$), which we define as the proportion of instances where the LLM ask for support:
\[B = \frac{N_{\mathrm{ask}}}{N}\]
where $N_{\mathrm{ask}}$ is the number of instances where the LLM asks for support, and $N$ denotes the total number of instances in the test set.
The other dimension is the performance improvement ($\Delta$, Delta):
\[\Delta = \frac{1}{N} \sum_{i=1}^{N_{\mathrm{ask}}} (h(y_i, \hat{y}_{i,z}) - h(y_i, \hat{y}_i))\]
where $h(\cdot)$ is the evaluation function of a given task, which takes ground truth $y_i$ and model output $\hat{y}_i$ as arguments ($\hat{y}_{i,z}$ is an output with the help of $z$).
Inspired by the idea behind the ROC curve~\cite{majnik2013roc}, we illustrate this trade-off with a graph, where the performance curve is plotted by adjusting the threshold $\tau$ from high to low along the x-axis.
We refer to this curve as Delta-Burden Curve (DBC) (see the leftmost subplot of Figure~\ref{fig:curves}).

\subsection{Methods for Seeking Support}
We design a prompt template $p_{\mathrm{ask}}(\cdot)$ to enable LLMs to request support by $\hat{a} = s(f(p_{\mathrm{ask}}(w) \mid \theta))$.
Here, $w$ represents the textual information that the LLM $f$ uses to determine whether it needs to seek support, and $s$ is the scoring function that converts the probability distribution of output tokens into a confidence score $\hat{a} \in [0,1]$.
We propose methods with varying compositions of $w$ to explore the information LLMs require to achieve better trade-off under DBC. 
Note that $p_{\mathrm{ask}}$ remains the same across all methods to minimize prompt engineering.
An overview of these methods is shown in Figure~\ref{fig:showcase}.

\noindent \textbf{Direct Ask (DA)}: $w = (db, x)$, composed of database schema $db$ and user data requirement $x$.

\noindent \textbf{Write then Ask (WA)}: $w = (db, x, \hat{y})$, where the LLM generates the SQL code $\hat{y} = f(p(db, x) \mid \theta)$ first and then use this self-generated output as the additional information in $w$.

\noindent \textbf{Execute then Ask (EA)}: $w = (db, x, \hat{y}, \hat{r})$, where the execution results $\hat{r}$ is returned by the database by executing LLM-generated SQL $\hat{y}$.

\section{Experiments}
\subsection{Dataset}

We use BIRD~\cite{li2024can}, which includes human-annotated external knowledge that serves as $z$.
For example, $z$ might be domain-specific knowledge, such as how to calculate financial indicators from database values.
The instruction $x$ represents the users' data requirements, paired with the ground truth SQL $y$.
It uses \textbf{Execution Accuracy (EX)} as the evaluation metric, where $h(y_i, \hat{y}_i)$ is defined as $\mathbbm{1}(r_i = \hat{r}_i)$.
Here, $r_i$ is the SQL execution result of $y_i$, and $\hat{r}_i$ is the execution result of $\hat{y}_i$.
Simply put, EX is the proportion of testing instances where $r_i$ and $\hat{r}_i$ are identical.

\begin{table*}[t!]
  \centering
  \begin{tabular}{lccccccc}
    \toprule
    \textbf{Methods/LLMs}&\textbf{Wizard}&\textbf{Llama3}&\textbf{DPSeek}&\textbf{GPT-3.5}&\textbf{Mixtral}&\textbf{GPT-4t}&\textbf{GPT-4o}\\
    \midrule
    Random Baseline      & 0.5000        & 0.5000        & 0.5000        & 0.5000         & 0.5000         & 0.5000        & 0.5000        \\
    \midrule
    Direct Ask           & 0.4915        & 0.4834        & 0.4976        & 0.4390         &\underline{0.5301}&\underline{0.5758}&\underline{0.5479}\\
    Write then Ask       & 0.4759        & 0.4497        & 0.4857        & 0.4735         &\underline{0.5677}&\underline{0.5807}&\underline{0.5740}\\
    Execute then Ask   &\underline{\textbf{0.5096}}&\textbf{0.4987}&\underline{\textbf{0.5848}}&\underline{\textbf{0.6313}}&\underline{\textbf{0.6242}}&\underline{\textbf{0.6641}}&\underline{\textbf{0.5989}}\\
    \bottomrule
  \end{tabular}
  \caption{Area Under Delta-Burden Curve (AUDBC) across different methods and LLMs. Text in \textbf{bold} denotes the method with the best performance, while \underline{underlined} text means better than random (uniform sampling of $\hat{a} \in [0,1]$).}
  \label{tab:audbc}
\end{table*}

\begin{table*}[t!]
  \centering
  \begin{tabular}{lccccccc}
    \toprule
    \textbf{Support/LLMs}&\textbf{Wizard}&\textbf{Llama3}&\textbf{DPSeek}&\textbf{GPT-3.5}&\textbf{Mixtral}&\textbf{GPT-4t}&\textbf{GPT-4o}\\
    \midrule
    w/o user support     & 0.1721 & 0.1767        & 0.2360        & 0.3064         & 0.2419         & 0.3142       & 0.3096        \\
    w/ full user support & 0.2764 & 0.3475        & 0.4185        & 0.4668         & 0.4126         & 0.4889       & 0.5117        \\
    \bottomrule
  \end{tabular}
  \caption{Execution accuracy (EX) of different support levels. Full user support means $B = 1$ (see Section~\ref{sec:eval}).}
  \label{tab:support}
\end{table*}

\subsection{Implementation}
\label{sec:implementation}
For open-weight LLMs, we use \textit{WizardCoder-34B}~\cite{Luo2023WizardCoderEC}, \textit{Llama-3-70b-chat}, \textit{DeepSeek-Coder-33B}~\cite{Guo2024DeepSeekCoderWT}, and \textit{Mixtral-8x22B}~\cite{Jiang2024MixtralOE} for diversity of different LLM families.
For closed-source LLMs, we use \textit{gpt-3.5-turbo-0125}, \textit{gpt-4-turbo-2024-04-09}, and \textit{gpt-4o-2024-05-13}~\cite{Achiam2023GPT4TR}.
The prompt $p_{\mathrm{ask}}(w)$ (included in Appendix~\ref{app:prompting}) instructs the model to output a single token \texttt{Yes/No} to indicate whether it needs support.
We define the scoring function $s$ as the softmax of \texttt{Yes} over log probabilities of \texttt{Yes} and \texttt{No} to derive $\hat{a} \in [0,1]$.

\section{Main Results}
\begin{figure*}[ht]
  \includegraphics[width=\linewidth]{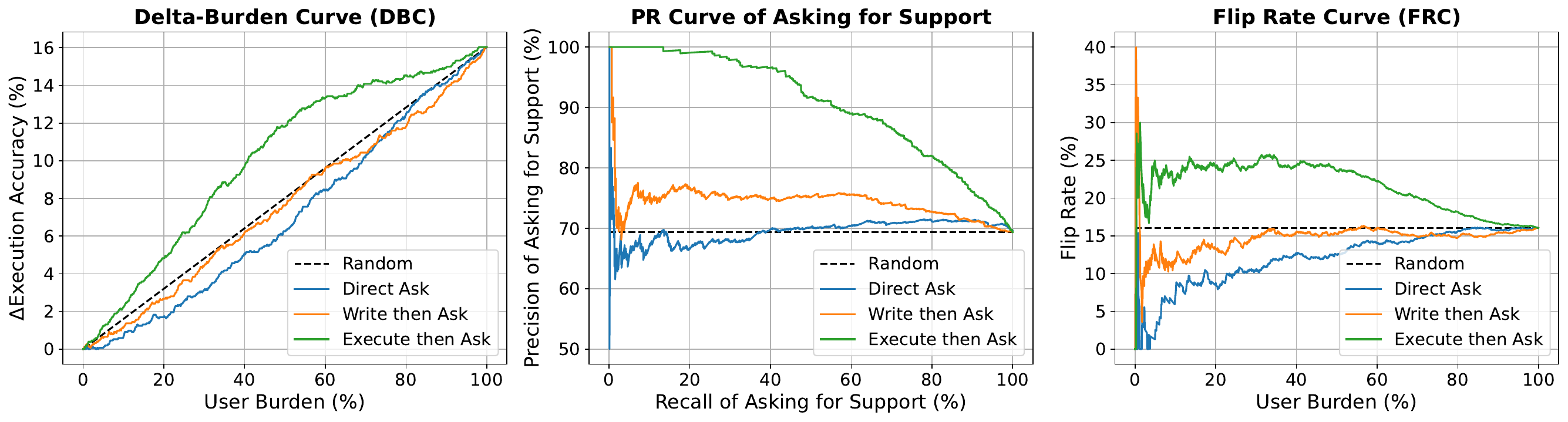}
  \caption{Performance curves of \textit{gpt-3.5-turbo-0125}. Curves of other LLMs are shown in Appendix~\ref{app:curves}.}
  \label{fig:curves}
\end{figure*}

Using the formulation in Section~\ref{sec:eval}, we quantify the performance of different methods with the Area Under Delta-Burden Curve (AUDBC) in Table~\ref{tab:audbc}.
Visualized DBCs are available in the leftmost subplots in Figure~\ref{fig:curves}.
Note that AUDBC should only be compared between methods under the same LLM, as it is normalized to the range of $[0,1]$ by dividing the area under the curve by the maximum square area, which depends on the scale of $\Delta$EX and differs across LLMs, as shown in Table~\ref{tab:support}.

There are three major findings:
(1) Execution then Ask consistently improves the performance-burden trade-off for LLMs, although \textit{Llama-3-70b-chat} fails to outperform the random baseline.
(2) The leftmost four LLMs in Table~\ref{tab:audbc} do not surpass the random baseline without the assistance of $\hat{r}$, indicating that many current LLMs still struggle to determine the need for support based on $x$ and $\hat{y}$ alone.
(3) Despite this, the rightmost three LLMs outperform the random baseline with the Write then Ask $(x, \hat{y})$ or even Direct Ask $(x)$ methods.
Nevertheless, the inclusion of $\hat{r}$ remains beneficial for further enhancing the trade-off between performance improvement and user burden.
Practical implications of the third point include the potential for cost savings by trading off the execution of $\hat{y}$ to obtain $\hat{r}$ in certain resource-constrained scenarios.

\section{Discussion}

\begin{table*}[t!]
  \centering
  \small
  \begin{tabular}{lccccccccc}
    \toprule
    \textbf{Methods/LLMs}&\textbf{Wizard}&\textbf{Llama3}&\textbf{DPSeek}&\textbf{GPT-3.5}&\textbf{Mixtral}&\textbf{GPT-4t}&\textbf{GPT-4o}&\textbf{Gemini}&\textbf{Claude}\\
    \midrule
    Random Baseline      & 0.5000        & 0.5000        & 0.5000        & 0.5000         & 0.5000         & 0.5000        & 0.5000 & 0.5000 & 0.5000        \\
    \midrule
    EA (real logprobs) &\underline{\textbf{0.5096}}&0.4987&\underline{\textbf{0.5848}}&\underline{\textbf{0.6313}}&\underline{\textbf{0.6242}}&\underline{\textbf{0.6641}}&\underline{\textbf{0.5989}}  & - & -      \\
    EA (verbalized) & \underline{0.5011}& \underline{\bf0.5333} & 0.4964 & \underline{0.5945} & \underline{0.6226} & 0.4850 & \underline{0.5152} & \underline{\bf0.5624} & \underline{\bf0.6174}\\
    \bottomrule
  \end{tabular}
  \caption{Area Under Delta-Burden Curve (AUDBC) with the verbalized token log probabilities approach. Text in \textbf{bold} denotes the method with the best performance, while \underline{underlined} text means better than random.}
  \label{tab:verb}
\end{table*}

\subsection{Analysis on the Delta-Burden Curves}
The Delta-Burden Curves (DBCs) plotted in Figure~\ref{fig:curves} quantify the following practical question: \textit{Under the same user burden, which method can achieve more performance boost?} To further analyze how this performance boost is achieved, we decompose the concept into two abilities:

\begin{enumerate}
    \item The ability to ask for support when the LLM cannot satisfy the instruction originally.
    \item The ability to utilize support effectively to \textit{flip} the incorrect output to the correct output.
\end{enumerate}

\noindent 1. For the first ability, we introduce the following metrics inspired by the precision-recall trade-off:

\noindent \textbf{Precision of Asking for Support} ($P_{\mathrm{ask}}$)
When the LLM asks for support, it should be the case that the LLM cannot satisfy the instruction originally, or it would cause unnecessary user burden:
\[P_{\mathrm{ask}} = \frac{\#(\text{AskforSupport} \And \text{OriginallyWrong})}{\#\text{AskforSupport}}\]

\noindent \textbf{Recall of Asking for Support} ($R_{\mathrm{ask}}$)
When the LLM is not able to satisfy the instruction originally, it should identify this need and ask for support:
\[R_{\mathrm{ask}} = \frac{\#(\text{AskforSupport} \And \text{OriginallyWrong})}{\#\text{OriginallyWrong}}\]

\noindent \textbf{PR Curve of Asking for Support}
Similar to how DBC is plotted, one can also adjust the threshold $\tau \in [0,1]$ from high to low along the x-axis to plot the Precision-Recall Curve of Asking for Support.

\noindent 2. For the second ability, we introduce \textit{Flip Rate}:

\noindent \textbf{Flip Rate}: This metric is calculated as the proportion of instances where the LLM's initially incorrect answers were corrected after receiving support, divided by the total number of instances where support was requested. Formally, it is defined as:
\[FR = \frac{1}{N_{ask}} \sum_{i=1}^{N_{\mathrm{ask}}} (h(y_i, \hat{y}_{i,z}) - h(y_i, \hat{y}_i))\]

\noindent Different from $\Delta$ defined in Section~\ref{sec:eval}, this metric emphasizes the \textit{efficiency} of leveraging support instead of the total improvement on the test set.
Like DBC, one may adjust the threshold $\tau$ to plot the Flip Rate Curve (FRC).
With the definition of these two abilities, we plot the DBC, PR Curve, and FRC on Figure~\ref{fig:curves}.
Although the Write then Ask method shows near-random performance in DBC, the PR Curve indicates it achieves better-than-random performance in identifying when support is needed. However, its lower Flip Rate suggests it is less efficient in utilizing the support to correct mistakes.
These two abilities, represented by the PR Curve and FRC, respectively, balance each other out, resulting in near-random performance on the DBC.
This finding shows that the ability to identify the need for support and the ability to utilize that support are distinct.
In future work, it is worth exploring how to further enhance each of these abilities.

\subsection{LLMs without Access to Log Probabilities}
Given that not all LLMs provide access to token log probabilities, we discuss how our method can be adapted for these ``black-box'' models. 
We modify the prompt template $p_{\mathrm{ask}}$ to $p_{\mathrm{verb}}$, which instructs the LLM to output the \textit{verbalized} confidence score $\hat{a}$ directly by specifying the range and meaning of $\hat{a} \in [0,1]$ in $p_{\mathrm{verb}}$ (attached in Appendix~\ref{app:verbal_prompt}).
In addition to the seven LLMs mentioned in Section~\ref{sec:implementation}, we also include two black-box models: \textit{gemini-1.0-pro-001} and \textit{claude-3-haiku-20240307}. The results, shown in Table~\ref{tab:verb}, indicate that using verbalized confidence scores generally degrades performance for most LLMs. However, it remains a promising alternative for black-box LLMs such as Gemini and Claude to surpass the random baseline.

\section{Related Work}
The ability of LLMs to identify the need for support relies on their \textit{well-calibratedness}~\cite{kadavath2022language}, which refers to their capacity to recognize uncertainty.
Previous studies focus on enhancing the calibration of predictions~\cite{xiao2022uncertainty, kuhn2023semantic}, or using verbalized token probabilities to achieve better calibration~\cite{tian2023just}.
Our work extends this line of research by exploring how LLMs can effectively seek user support by leveraging their well-calibrated property.
The major distinction between this and existing calibration studies lies in extending the focus from \textit{identifying} the uncertainty to \textit{utilizing} support.

\section{Conclusion}
We propose a framework for LLMs to seek support, and evaluate methods on Text-to-SQL generation.
Our findings suggest the importance of external signals, such as SQL execution results, in helping LLMs better manage performance-burden trade-off.
We further decompose DBC into the ability of \textit{identify} the need for support and the ability to \textit{utilize} the support.
Future works may explore a broader range of tasks or develop methods to improve both the identification and utilization of support.

\pagebreak

\section{Limitations}
\subsection{Task Coverage}

The scope of our experiments is limited to the Text-to-SQL task. While this task provides a useful case study for evaluating LLMs' ability to seek and utilize support, it does not encompass the full range of potential applications for LLMs. Future work should extend the evaluation to a broader set of tasks to ensure the generalizability of our findings.

\subsection{Types of Support}

In this study, we primarily focus on a single type of support: human-annotated external knowledge.
However, there are many other types of support that LLMs might require.
Future works could explore how LLMs can request and utilize these various forms of support to enhance their performance.

\subsection{Dependence on External Feedback}

Our findings indicate that LLMs significantly benefit from external signals, such as SQL execution results. However, this reliance on external feedback may not always be feasible in practical applications, where immediate execution or access to external data might be limited. Developing methods that enable LLMs to better manage without such feedback remains an important area for future exploration.


\begin{thebibliography}{13}
\providecommand{\natexlab}[1]{#1}

\bibitem[{Guo et~al.(2024)Guo, Zhu, Yang, Xie, Dong, Zhang, Chen, Bi, Wu, Li, Luo, Xiong, and Liang}]{Guo2024DeepSeekCoderWT}
Daya Guo, Qihao Zhu, Dejian Yang, Zhenda Xie, Kai Dong, Wentao Zhang, Guanting Chen, Xiao Bi, Yu~Wu, Y.~K. Li, Fuli Luo, Yingfei Xiong, and Wenfeng Liang. 2024.
\newblock Deepseek-coder: When the large language model meets programming - the rise of code intelligence.
\newblock \emph{ArXiv}, abs/2401.14196.

\bibitem[{Jiang et~al.(2024)Jiang, Sablayrolles, Roux, Mensch, Savary, Bamford, Chaplot, de~Las~Casas, Hanna, Bressand, Lengyel, Bour, Lample, Lavaud, Saulnier, Lachaux, Stock, Subramanian, Yang, Antoniak, Scao, Gervet, Lavril, Wang, Lacroix, and Sayed}]{Jiang2024MixtralOE}
Albert~Q. Jiang, Alexandre Sablayrolles, Antoine Roux, Arthur Mensch, Blanche Savary, Chris Bamford, Devendra~Singh Chaplot, Diego de~Las~Casas, Emma~Bou Hanna, Florian Bressand, Gianna Lengyel, Guillaume Bour, Guillaume Lample, L'elio~Renard Lavaud, Lucile Saulnier, Marie-Anne Lachaux, Pierre Stock, Sandeep Subramanian, Sophia Yang, Szymon Antoniak, Teven~Le Scao, Th{\'e}ophile Gervet, Thibaut Lavril, Thomas Wang, Timoth{\'e}e Lacroix, and William~El Sayed. 2024.
\newblock Mixtral of experts.
\newblock \emph{ArXiv}, abs/2401.04088.

\bibitem[{Kadavath et~al.(2022)Kadavath, Conerly, Askell, Henighan, Drain, Perez, Schiefer, Hatfield-Dodds, DasSarma, Tran-Johnson et~al.}]{kadavath2022language}
Saurav Kadavath, Tom Conerly, Amanda Askell, Tom Henighan, Dawn Drain, Ethan Perez, Nicholas Schiefer, Zac Hatfield-Dodds, Nova DasSarma, Eli Tran-Johnson, et~al. 2022.
\newblock Language models (mostly) know what they know.
\newblock \emph{arXiv preprint arXiv:2207.05221}.

\bibitem[{Kuhn et~al.(2023)Kuhn, Gal, and Farquhar}]{kuhn2023semantic}
Lorenz Kuhn, Yarin Gal, and Sebastian Farquhar. 2023.
\newblock Semantic uncertainty: Linguistic invariances for uncertainty estimation in natural language generation.
\newblock \emph{arXiv preprint arXiv:2302.09664}.

\bibitem[{Li et~al.(2024)Li, Hui, Qu, Yang, Li, Li, Wang, Qin, Geng, Huo et~al.}]{li2024can}
Jinyang Li, Binyuan Hui, Ge~Qu, Jiaxi Yang, Binhua Li, Bowen Li, Bailin Wang, Bowen Qin, Ruiying Geng, Nan Huo, et~al. 2024.
\newblock Can llm already serve as a database interface? a big bench for large-scale database grounded text-to-sqls.
\newblock \emph{Advances in Neural Information Processing Systems}, 36.

\bibitem[{Luo et~al.(2023)Luo, Xu, Zhao, Sun, Geng, Hu, Tao, Ma, Lin, and Jiang}]{Luo2023WizardCoderEC}
Ziyang Luo, Can Xu, Pu~Zhao, Qingfeng Sun, Xiubo Geng, Wenxiang Hu, Chongyang Tao, Jing Ma, Qingwei Lin, and Daxin Jiang. 2023.
\newblock Wizardcoder: Empowering code large language models with evol-instruct.
\newblock \emph{ArXiv}, abs/2306.08568.

\bibitem[{Majnik and Bosni{\'c}(2013)}]{majnik2013roc}
Matja{\v{z}} Majnik and Zoran Bosni{\'c}. 2013.
\newblock Roc analysis of classifiers in machine learning: A survey.
\newblock \emph{Intelligent data analysis}, 17(3):531--558.

\bibitem[{OpenAI(2023)}]{Achiam2023GPT4TR}
OpenAI. 2023.
\newblock Gpt-4 technical report.

\bibitem[{Pourreza and Rafiei(2023)}]{pourreza2023evaluating}
Mohammadreza Pourreza and Davood Rafiei. 2023.
\newblock Evaluating cross-domain text-to-sql models and benchmarks.
\newblock \emph{arXiv preprint arXiv:2310.18538}.

\bibitem[{Rawte et~al.(2023)Rawte, Sheth, and Das}]{rawte2023survey}
Vipula Rawte, Amit Sheth, and Amitava Das. 2023.
\newblock A survey of hallucination in large foundation models.
\newblock \emph{arXiv preprint arXiv:2309.05922}.

\bibitem[{Tian et~al.(2023)Tian, Mitchell, Zhou, Sharma, Rafailov, Yao, Finn, and Manning}]{tian2023just}
Katherine Tian, Eric Mitchell, Allan Zhou, Archit Sharma, Rafael Rafailov, Huaxiu Yao, Chelsea Finn, and Christopher~D Manning. 2023.
\newblock Just ask for calibration: Strategies for eliciting calibrated confidence scores from language models fine-tuned with human feedback.
\newblock In \emph{Proceedings of the 2023 Conference on Empirical Methods in Natural Language Processing}, pages 5433--5442.

\bibitem[{Wei et~al.(2021)Wei, Bosma, Zhao, Guu, Yu, Lester, Du, Dai, and Le}]{wei2021finetuned}
Jason Wei, Maarten Bosma, Vincent~Y Zhao, Kelvin Guu, Adams~Wei Yu, Brian Lester, Nan Du, Andrew~M Dai, and Quoc~V Le. 2021.
\newblock Finetuned language models are zero-shot learners.
\newblock \emph{arXiv preprint arXiv:2109.01652}.

\bibitem[{Xiao et~al.(2022)Xiao, Liang, Bhatt, Neiswanger, Salakhutdinov, and Morency}]{xiao2022uncertainty}
Yuxin Xiao, Paul~Pu Liang, Umang Bhatt, Willie Neiswanger, Ruslan Salakhutdinov, and Louis-Philippe Morency. 2022.
\newblock Uncertainty quantification with pre-trained language models: A large-scale empirical analysis.
\newblock \emph{arXiv preprint arXiv:2210.04714}.

\end{thebibliography}

\pagebreak

\appendix

\section{Prompt Templates}
\label{app:prompting}

We include the prompt templates used in this work.

\subsection{Prompt for Seeking Support}
The prompt template $p_{\mathrm{ask}}(w)$ used to instruct LLMs for seeking support is as follows:

\noindent
\begin{minipage}{\columnwidth}
    \begin{mdframed}
    You are currently doing the text-to-SQL task. Based on the information provided (\textcolor{blue}{\{items\}}), you have to determine whether additional hints are required for you to generate the SQL correctly to answer the user's question. You should only ask for additional hints when you actually need them, since you will also be evaluated based on the number of times you ask for hints, which would be provided by the user.\\
    \\
    information provided (enclosed by triple backticks): \\
    ```\\
    \textcolor{blue}{\{information\}}\\
    ```\\
    \\
    Answer a single word Yes if you need hints (since the information provided is not enough to generate SQL correctly). Answer a single word No if hints are not required (since you are already confident to generate SQL).\\
    Do you need additional hints? Answer (Yes / No):
    \end{mdframed}
    \label{fig:prompt_support}
\end{minipage}

In this template, the actual contents of \textcolor{blue}{\{items\}} and \textcolor{blue}{\{information\}} depend on the method used.
The contents are summarized in Table~\ref{tab:support_contents}.
For example, $w = (db, x, \hat{y}, \hat{r})$ in Execute then Ask (EA), so \textcolor{blue}{\{items\}} will be filled with the four item names and \textcolor{blue}{\{information\}} will be replaced by actual information of the four items.
Similarly for Write then Ask ($w = (db, x, \hat{y})$) and Direct Ask ($w = (db, x)$).
\begin{table}[h!]
  \centering
  \begin{tabular}{lcc}
    \toprule
    \textbf{Item} & \textbf{Item Name} & \textbf{Information} \\
    \midrule
    \textbf{$db$} & Database schema & \textcolor{red}{\{db\_schema\}} \\
    \textbf{$x$} & User's question & \textcolor{red}{\{question\}} \\
    \textbf{$\hat{y}$} & Generated SQL & \textcolor{red}{\{gen\_sql\}} \\
    \textbf{$\hat{r}$} & SQL execution results & \textcolor{red}{\{exe\_results\}} \\
    \bottomrule
  \end{tabular}
  \caption{Contents in the prompt $p_{\mathrm{ask}}(w)$, where \textcolor{blue}{\{items\}} will be filled with words in the ``Item Name'' column, while \textcolor{blue}{\{information\}} is replaced with actual information of text in \textcolor{red}{\{red\}}.}
  \label{tab:support_contents}
\end{table}


\subsection{Prompt for Seeking Support (Verbalized)}
\label{app:verbal_prompt}
The prompt template for generating verbalized probabilities in LLMs without access to token log probabilities (e.g., Gemini and Claude families):

\noindent
\begin{minipage}{\columnwidth}
    \begin{mdframed}
    You are currently doing the text-to-SQL task. Based on the information provided (\textcolor{blue}{\{items\}}), you have to determine whether additional hints are required for you to generate the SQL correctly to answer the user's question. You should only ask for additional hints when you actually need them, since you will also be evaluated based on the number of times you ask for hints, which would be provided by the user.\\
    \\
    information provided (enclosed by triple backticks): \\
    ```\\
    \textcolor{blue}{\{information\}}\\
    ```\\
    \\
    Do you need additional hints? Provide the precise probability that you need hints (closer to 0 means you don't need hints, closer to 1 means you need hints).\\
    Give ONLY the precise probability to five decimal places (format: 0.abcde, where abcde can be different digits), no other words or explanations are needed.
    \end{mdframed}
    \label{fig:prompt_verbalized}
\end{minipage}

The prompt template is similar to the original template shown in~\ref{fig:prompt_support}, except that the last few sentences are modified.

\vfill\null

\subsection{Prompt for Generating SQL Code}
The prompt template $p(\cdot)$ for converting user data requirement $x$ into SQL code is as follows:

\noindent
\begin{minipage}{\columnwidth}
    \begin{mdframed}
    \textcolor{blue}{\{db\_schema\}}\\
    \\
    -- Using valid SQLite, answer the following questions for the tables provided above.\\
    -- Question: \textcolor{blue}{\{question\}}\\
    Now, generate the correct SQL code directly in the format of ```sql\textbackslash n<your\_SQL\_code>\textbackslash n```:
    \end{mdframed}
    \label{fig:prompt_sql}
\end{minipage}

If user support $z$ is provided (i.e., when LLMs ask for support), the prompt template is slightly modified as follows:

\noindent
\begin{minipage}{\columnwidth}
    \begin{mdframed}
    \textcolor{blue}{\{db\_schema\}}\\
    \\
    -- External Knowledge: \textcolor{blue}{\{support\}}\\
    -- Using valid SQLite, answer the following questions for the tables provided above. You can use the provided External Knowledge to help you generate valid and correct SQLite.\\
    -- Question: \textcolor{blue}{\{question\}}\\
    Now, generate the correct SQL code directly in the format of ```sql\textbackslash n<your\_SQL\_code>\textbackslash n```:
    \end{mdframed}
    \label{fig:prompt_sqlwithz}
\end{minipage}

In these two templates, \textcolor{blue}{\{db\_schema\}} is $db$, \textcolor{blue}{\{question\}} is user data requirement $x$, and \textcolor{blue}{\{support\}} is user support $z$, which is human-annotated external knowledge in BIRD~\cite{li2024can}.
\section{Performance Curves}
\label{app:curves}

\begin{figure*}[ht]
  \includegraphics[width=\linewidth]{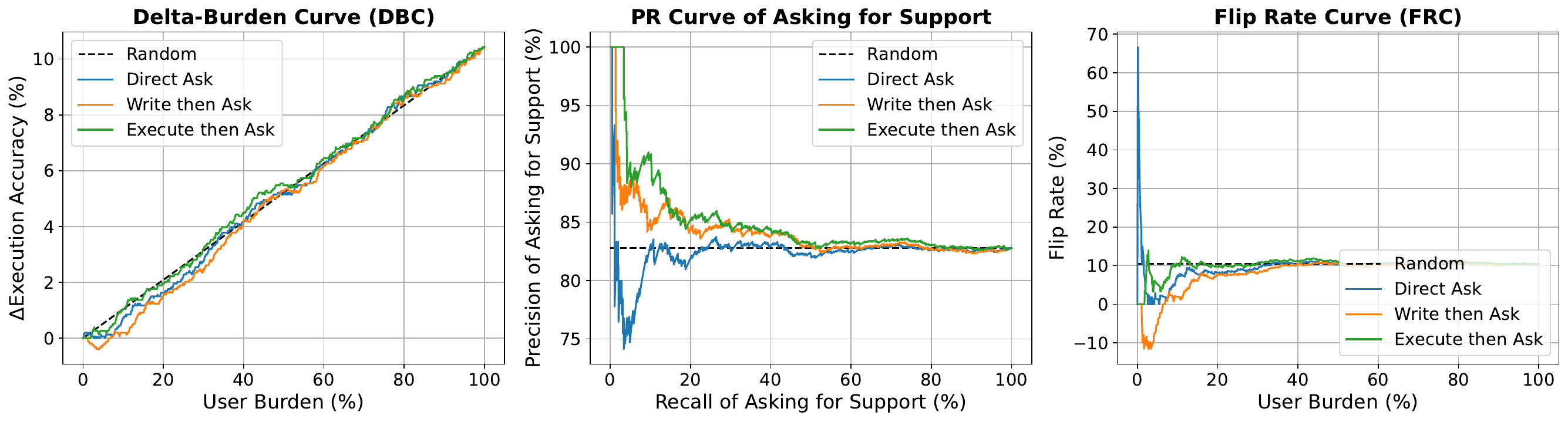}
  \caption{Performance curves of \textit{WizardCoder-Python-34B-V1.0}.}
  \label{fig:wizard_curves}
\end{figure*}

\begin{figure*}[ht]
  \includegraphics[width=\linewidth]{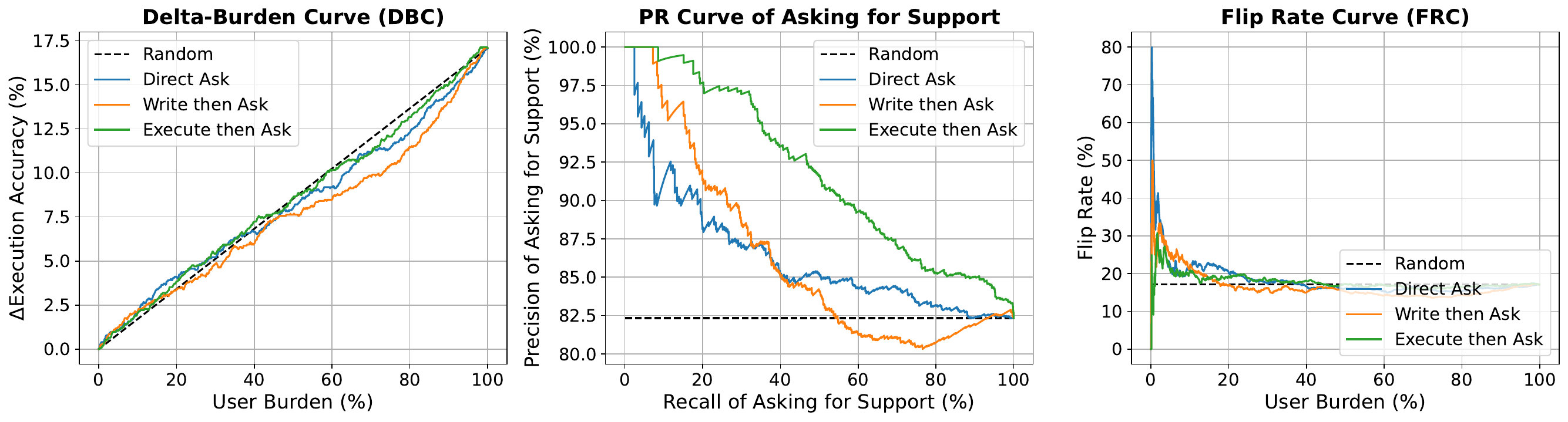}
  \caption{Performance curves of \textit{Llama-3-70b-chat-hf}.}
  \label{fig:llama3_curves}
\end{figure*}

\begin{figure*}[ht]
  \includegraphics[width=\linewidth]{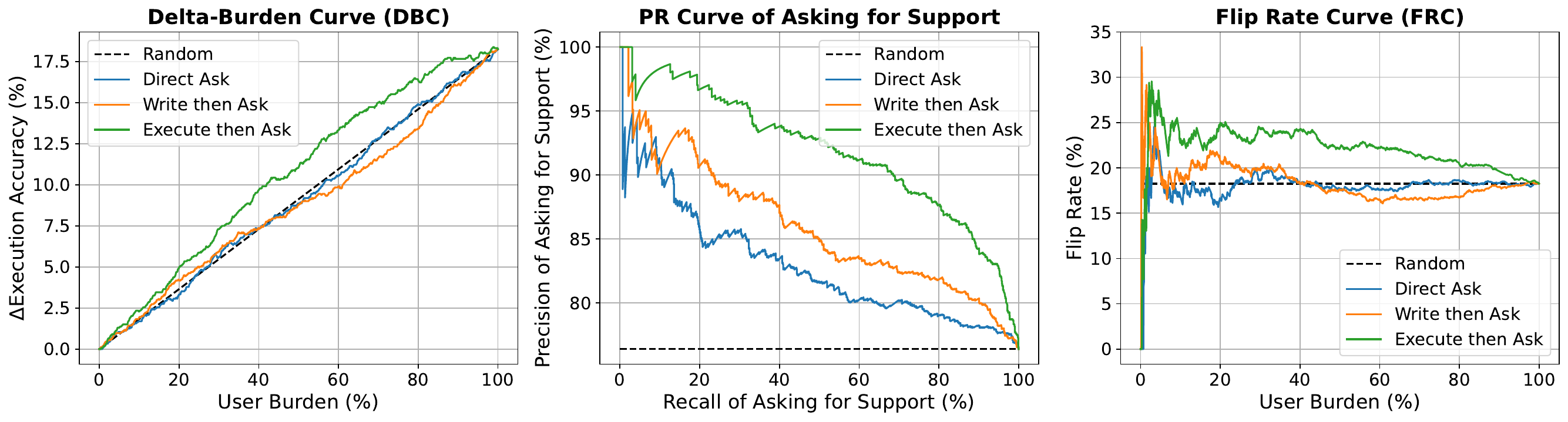}
  \caption{Performance curves of \textit{deepseek-coder-33b-instruct}.}
  \label{fig:deepseek_curves}
\end{figure*}

\begin{figure*}[ht]
  \includegraphics[width=\linewidth]{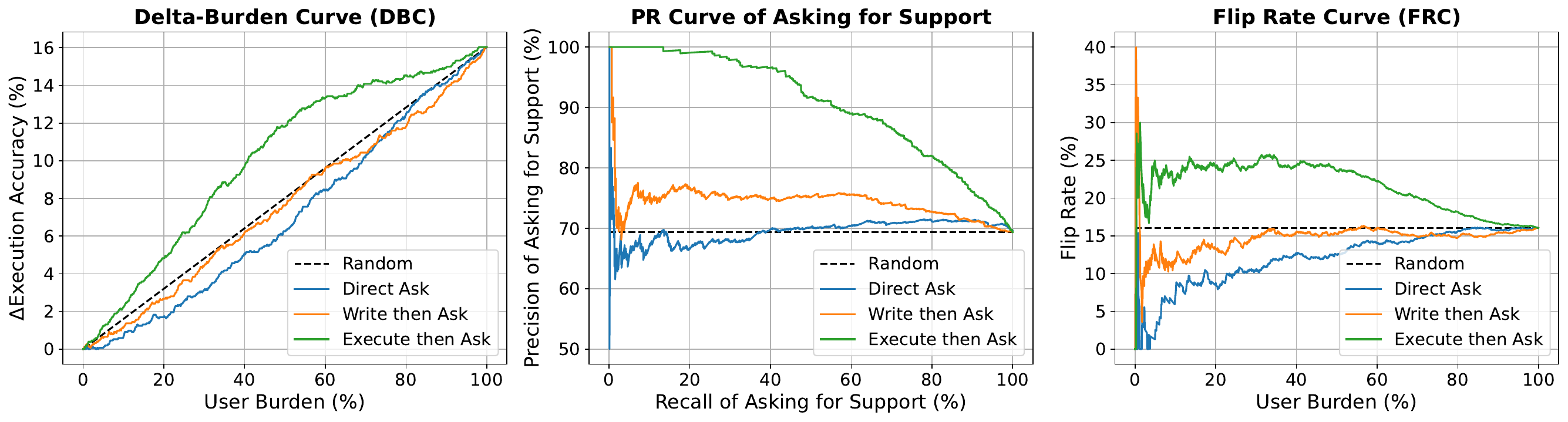}
  \caption{Performance curves of \textit{gpt-3.5-turbo-0125}.}
  \label{fig:gpt35_curves}
\end{figure*}

\begin{figure*}[ht]
  \includegraphics[width=\linewidth]{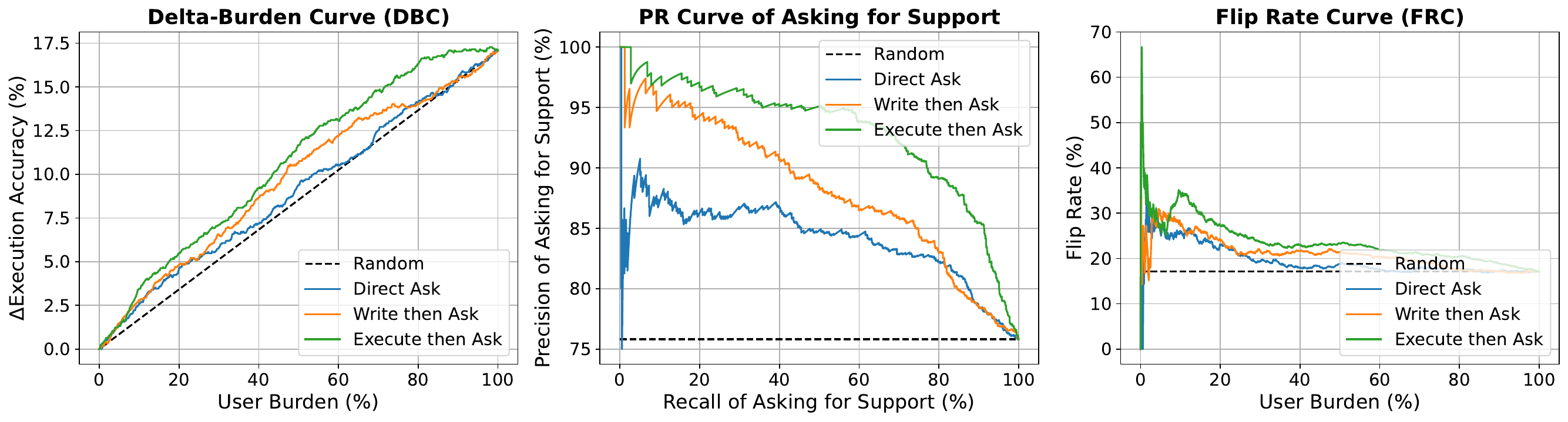}
  \caption{Performance curves of \textit{Mixtral-8x22B-Instruct-v0.1}.}
  \label{fig:mixtral_curves}
\end{figure*}

\begin{figure*}[ht]
  \includegraphics[width=\linewidth]{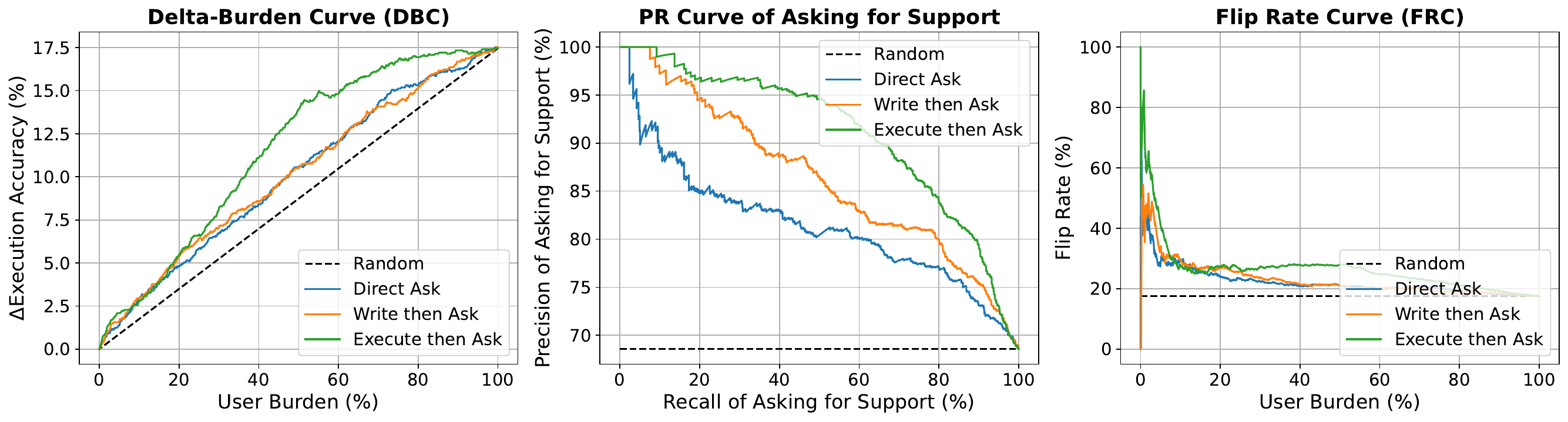}
  \caption{Performance curves of \textit{gpt-4-turbo-2024-04-09}.}
  \label{fig:gpt4t_curves}
\end{figure*}

\begin{figure*}[ht]
  \includegraphics[width=\linewidth]{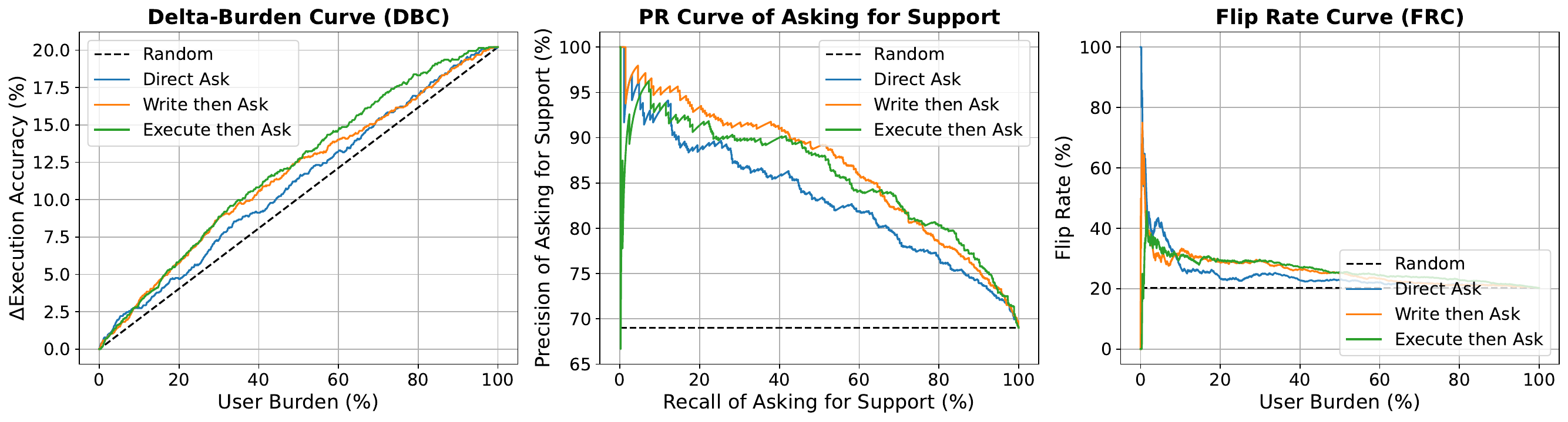}
  \caption{Performance curves of \textit{gpt-4o-2024-05-13}.}
  \label{fig:gpt_4o_curves}
\end{figure*}

We present visualizations of all performance curves in Table~\ref{fig:wizard_curves}, \ref{fig:llama3_curves}, \ref{fig:deepseek_curves}, \ref{fig:gpt35_curves}, \ref{fig:mixtral_curves}, \ref{fig:gpt4t_curves}, and~\ref{fig:gpt_4o_curves}.

\end{document}